\documentclass[utf8]{frontiersSCNS} 

\usepackage{url,hyperref,lineno,microtype,subcaption}
\usepackage[onehalfspacing]{setspace}
\usepackage{graphicx}
\usepackage[ruled,vlined,linesnumbered,oldcommands]{algorithm2e}
\usepackage{cite}
\usepackage{comment}

\usepackage{caption}
\usepackage{subcaption}

\usepackage{multirow}

\usepackage{soul}
\usepackage{xcolor}
\usepackage{xparse}
\makeatletter

\ExplSyntaxOn
\cs_new:Npn \white_text:n #1
  {
    \fp_set:Nn \l_tmpa_fp {#1 * .01}
    \llap{\textcolor{white}{\the\SOUL@syllable}\hspace{\fp_to_decimal:N \l_tmpa_fp em}}
    \llap{\textcolor{white}{\the\SOUL@syllable}\hspace{-\fp_to_decimal:N \l_tmpa_fp em}}
  }
\NewDocumentCommand{\whiten}{ m }
    {
      \int_step_function:nnnN {1}{1}{#1} \white_text:n
    }
\ExplSyntaxOff

\NewDocumentCommand{ \fpul }{ D<>{5} O{0.2ex} O{0.1ex} +m } {%
\begingroup
\setul{#2}{#3}%
\def\SOUL@uleverysyllable{%
   \setbox0=\hbox{\the\SOUL@syllable}%
   \ifdim\dp0>\z@
      \SOUL@ulunderline{\phantom{\the\SOUL@syllable}}%
      \whiten{#1}%
      \llap{%
        \the\SOUL@syllable
        \SOUL@setkern\SOUL@charkern
      }%
   \else
       \SOUL@ulunderline{%
         \the\SOUL@syllable
         \SOUL@setkern\SOUL@charkern
       }%
   \fi}%
    \ul{#4}%
\endgroup
}

\makeatother

\makeatletter
\def\tok@scan#1{%
  \ifx#1\relax
    \let\tok@next\relax
  \else
    \edef\my@list{\my@list#1}%
    \let\tok@next\tok@scan
  \fi
  \tok@next
}
\newcommand{\@strip}[2]{%
  \def\my@list{}\tok@scan#2\relax\let#1\my@list}
\newcommand{\Cite}[1]{\@strip\@args{#1}\cite\@args}
\makeatother

\DeclareMathOperator{\uir}{HIRR}

\DeclareMathOperator{\eqm}{EqM}
\DeclareMathOperator{\opp}{Opp}
\DeclareMathOperator{\opps}{Opps}
\DeclareMathOperator{\oppdeg}{oppdeg}
\DeclareMathOperator{\eq}{Eq}
\DeclareMathOperator{\des}{Des}
\DeclareMathOperator{\desp}{Des'}
\DeclareMathOperator{\bnf}{Bnf}
\DeclareMathOperator{\bnfp}{Bnf'}
\DeclareMathOperator*{\argmin}{argmin}

\DeclareMathOperator{\dom}{dom}
\def\tmin#1{\ensuremath{\inf_{#1}}}
\def\tmax#1{\ensuremath{\sup_{#1}}}
\def\ddes{\mu_{\des}}

\newcommand{\free}[1]{F^{#1}}
\newcommand{\pset}{{\mathcal{P}}}
\newcommand{\psetp}{\mathcal{P}^{+}}
\newcommand{\lang}{\mathcal{L}}
\newcommand{\inte}{\mathfrak{i}}
\newcommand{\ints}{\mathfrak{I}}

\def\keyFont{\fontsize{8}{11}\helveticabold }
\def\firstAuthorLast{Buyukgoz {et~al.}} 
\def\Authors{Sera Buyukgoz\,$^{1,2,\dagger}$, Jasmin Grosinger\,$^{3,\dagger}$, Mohamed Chetouani\,$^{2,*}$,  and Alessandro Saffiotti\,$^{3}$}
\def\Address{$^{1}$SoftBank Robotics Europe, Paris, France \\
$^{2}$Sorbonne  University,  Institute  for  Intelligent  Systems  and  Robotics,  CNRS  UMR  7222, Paris,  France \\
$^{3}$AASS Cognitive Robotic Systems Lab
School of Science and Technology, Orebro University, Orebro, Sweden \\
$^{\dagger}$These authors have contributed equally to this work and share first authorship}
\def\corrAuthor{Mohamed Chetouani, Sorbonne  University,  4 Place Jussieu
75252 Paris Cedex 05, France}

\def\corrEmail{mohamed.chetouani@sorbonne-universite.fr}

\definecolor{myblue}{rgb}{0,0,0.7}

\definecolor{myred}{rgb}{0.7,0,0}

\definecolor{mygreen}{rgb}{0,0.4,0}

\definecolor{jgblue}{rgb}{0.3,0.25,0.95}

\definecolor{mypurple}{rgb}{0.4,0.0,0.5}

\definecolor{llgray}{rgb}{0.86, 0.86, 0.86}

\begin{document}
\onecolumn
\firstpage{1}




\title{Two ways to make your robot proactive: reasoning about human
  intentions, or reasoning about possible futures}



\author[\firstAuthorLast ]{\Authors} 
\address{} 
\correspondance{} 

\extraAuth{}

\maketitle
\begin{abstract}
Robots sharing their space with humans need to be proactive in order to
be helpful. Proactive robots are able to act on their own initiative in
an anticipatory way to benefit humans. In this work, we investigate two
ways to make robots proactive.  One way is to recognize human's
intentions and to act to fulfill them, like opening the door that you
are about to cross.  The other way is to reason about possible future
threats or opportunities and to act to prevent or to foster them, like
recommending you to take an umbrella since rain has been forcasted.  In
this paper, we present approaches to realize these two types of
proactive behavior.  We then present an integrated system that can
generate proactive robot behavior by reasoning on both factors:
intentions and predictions.  We illustrate our system on a sample use
case including a domestic robot and a human.  We first run this use case
with the two separate proactive systems, intention-based and
prediction-based, and then run it with our integrated system.  The
results show that the integrated system is able to take into account a broader variety of aspects that are needed for proactivity.

\tiny \keyFont{ \section{Keywords:} Proactive agents, Human intentions,
  Autonomous robots, Social robot, Human-centered AI, Human-Robot
  interaction}


\end{abstract}

\section{Introduction}

Humans can act on their own initiative.
%
Imagine the following scenario: you see your flatmate preparing to leave
for a hiking trip in a rainy zone.  It is quite likely that you will
give your flatmate some advice like checking the weather forecast or
taking some extra equipment.  Such behavior even occurs between
strangers. When people see a person holding garbage and looking around,
they tend to show where the garbage bin is since they recognized the
person's intention to dispose of their garbage.  This type of intuitive
interaction is common between humans, and it is already observed in
infants~\citep{WarnekenTomasello.science2006}.
The question is, what happens if one of the actors is a robot?
The robot should be able to recognize and reason about the human's
intentions; to reason about the current and forecasted states of the
environment; to understand what states may be preferrable to others; and
to forsee problems that the human could face. The robot should also be
able to reason about the potential effects of its own actions, and
select and perform actions that support the human given this
context. The behavior of initiating own action taking into account all
these aspects is called \emph{proactive behavior}.

Most of the existing work in human-robot interaction rely on the human
taking the initiative: the human sets a request, and the robot generates
and executes a plan to satisfy it. However, in the above examples of
human-to-human interaction there is no explicit request.  The
interaction works because humans are able to assess other humans'
intentions, anticipate consequences, and reason about preferred states.
In this paper, we are interested in proactive human-robot interaction,
that is, interactions where the robot behaves by acting on its own
initiative, in an anticipatory way, and without being given an explicit
goal~\citep{grant2008the,peng2019design,grosinger2019robots}.  We
consider two types of proactive robot behavior: one in which the robot
understands the human's intentions and helps the human to achieve them; and
one in which the robot foresees possible future situations that are
undesirable (or desirable) according to the human's preferences, and
acts to avoid (or foster) them.


Specifically, we propose a framework that identifies opportunities for
acting and selects some of them for execution.  \emph{Opportunities}
here are formal concepts grounded in the relation between acting,
preferences and state predictions.
Our framework includes two main mechanisms that contribute to initiating
proactive behavior; \emph{human intention recognition and reasoning} and
\emph{equilibrium maintenance}. The former mechanism is based on
recognizing human intent from a known list of possible intents that the
human can have. The latter one is based on predicting how the state may
evolve in time and comparing preferences of states resulting from
different actions (or inaction) \citep{grosinger2019robots}.  These two
mechanisms correspond to the two types of proactive behavior mentioned
above: intention-based, and prediction-based.  The whole framework
includes provisions to combine these mechanisms into an integrated
proactive system.

The main contributions of this paper are: (i) we propose a novel method
based on human intention recognition to generate intention-based
proactive robot behavior, that we call \emph{human intention recognition
and reasoning ($\uir$)}; (ii) we adapt an existing method based on
temporal predictions and state preferences, called \emph{equilibrium
maintenance ($\eqm$)}, to generate prediction-based proactive robot
behavior; (iii) we define an architecture to combine both methods to
create a proactive robot that considers both human intentions and
temporal predictions; and (iv) we compare all these using a sample case
study.


The rest of this paper is organized as follows.  The next section
presents the necessary background together with related work on
intention recognition, proactivity, and their combination.  In
Section~\ref{sec:sys} we define our systems for intention-based
proactivity and for prediction-based proactivity, together with their
integration.  Section~\ref{sec:exp} describes the implementation and
shows the results of a task involving a simulated domestic robot and a
human.  Finally, in Section~\ref{sec:disc} we discuss our results and
conclude.

\section{Background and Related Work}\label{sec:sota}


In our work, we combine intention recognition and temporal predictions
to generate proactive behavior.  Here we provide the relevant background
and related work on these research areas.


\subsection{Intention Recognition}

In order to assist humans, a robot requires some knowledge of the
human's goals and intentions.
In Belief-Desire-Intention (BDI) models, 
\citep{rao1995bdi}, the agent represents the environment in terms of
beliefs that are true.
A set of desires, representing the agent's goals, guides the agent's
behavior. We may or may not know the agent's goals. The intention
represents the path that the agent is currently taking to reach a goal. \citet{Bratman1989} points out that the concept of intention is used to
characterize both the human's actions and mind (mental states). Actions
are considered as done with a certain intention. Humans attribute mental
states of intending to other agents such as having an intention to act
in certain ways now or later. In this paper, we consider an intention to be a mental state that is expressed through goal-directed actions.

{\it Intention recognition} is the process of inferring an agent's intention by analyzing their actions and their actions' effects on the environment
\citep{Han2013StateMaking}. Approaches in action recognition, goal
recognition and plan recognition have been used to infer intention. According to \citet{vanhorenbeke2021recognition}, intention recognition
systems can be classified as logic-based, classical machine learning,
deep learning and brain-inspired approaches, or they can be classified
in terms of the behavior of the observed human towards the observer. We take here a simplified view, and consider two classes of intention
recognition approaches: logic-based and probabilistic.
%
Logic-based approaches are defined by a set of domain-independent rules
that capture the relevant knowledge to infer the human's intention
through deduction \citep{sukthankar2014plan}. The knowledge can be
represented in different ways, including using plan representation languages
like STRIPS and PDDL that describe the state of the environment and the
effects of the possible actions.
%
Logic-based approaches work well in highly structured environments.
Logic representation can define different kinds of relationships
depending on the problem. These relationships allow to recognize humans'
intentions based on observations. Another advantage of logic-based
approaches is that they are highly expressive. The reasoning result can
potentially be traceable and human-understandable.  However, many
logic-based approaches assume that the human is rational and try to find
the optimal intention that best fits the observations, while humans
often act in non-rational ways \citep{Dreyfus2007}. This makes
logic-based approaches less reliable in real-world problems. The
uncertainty in humans' rationality might be addressed by a combination
of logic-based with probabilistic reasoning techniques.
%
Probabilistic approaches exploit Bayesian networks and Markov models.
Bayesian networks are generative probabilistic graphical models that
represent random variables as nodes and conditional dependencies as
arrows between them \citep{vanhorenbeke2021recognition}. They can
provide the probability distribution of any set of random variables
given another set of observed variables. Some planning systems use Bayesian inference to reason about
intention. Such approaches are referred to as Bayesian inverse
planning. \citet{ramirez2009Plan} propose an approximate planning method
that generates a set of possible goals by using Bayesian inverse
planning methods on classical planning. The method assumes that humans
are perfectly rational, which means they only optimally pursue their
goals. As a result of this, the indecisive behavior of humans is not
tolerated. This limitation is partly addressed in \citet{ramirez2010}, that
introduces a more general formulation.
\citet{persiani2020Computational} offers an example of using Bayesian
inference in a logic based approach. The authors use classical planning
to generate an action plan for each goal, then they use a Bayesian prior
function to infer human intention. Probabilistic approaches are able to handle uncertainty, and can
therefore handle real-world settings such as non-rational agents,
interrupted plans and partial observability
\citep{vanhorenbeke2021recognition}.  On the other hand, they are less
expressive than logic-based systems, since it is hard to understand the
reasoning behind the result.  Scalability is another, well-known
difficulty with probabilistic approaches.

In this work, we adopt a logic-based approach
as done in \citet{persiani2020Computational}: we represent the robot's
knowledge in a symbolic form, which the robot uses to plan its actions.
We assume rational humans. The approach gives us the advantage of
getting results that are easily traceable and human-readable.

\subsection{Proactivity}

\emph{Proactive} AI systems and robots are opposed to \emph{reactive} AI
systems, which respond to explicit requests or external events.  In
organizational psychology proactive behavior is understood as
\emph{anticipatory, self-initiated action}~\citep{grant2008the}.  When
it comes to artificial agents, though, we lack a clear definition of
proactivity.  Drawing inspiration from the human proactive process, we
can identify the functionalities that are needed for artificial
proactivity: context-awareness, activity recognition, goal reasoning,
planning, and plan execution and execution monitoring.  Each one of
these functionalities in itself has been the subject of active
research~\citep{doush2020survey, wang2019deep, aha2018goal,
  ghallab2016automated, hertzbergEtAl2016aireasoning}. Proactivity needs
to contemplate these areas jointly and in a separate process to each.
Context-awareness is not the central topic in proactivity, but it is
used to understand what the current situation is, and with this
knowledge it is possible to decide how to act. Goal Reasoning (GR) deals
with questions about generating, selecting, maintaining and dispatching
goals for execution~\citep{aha2018goal}. Planning can be described as searching and selecting an optimal action
trajectory to a goal that is given externally by the human or by some
trigger. Proactivity resides on the abstraction level above. It is
finding the acting decisions or goals that should be planned for, hence,
it produces the input to a planner.  Finally, plan execution and
monitoring are employed by Proactivity to enact the acting decision
inferred and to invoke new reasoning on proactivity when execution
fails.

Recently there has been a number of promising works in the field of
artificial proactivity. \citet{baraglia2017efficient} address the
question whether and when a robot should take initiative during joint
human-robot task execution. The domain used is table-top manipulation
tasks. \citet{baraglia2017efficient} employ dynamic Bayesian networks to
predict environmental states and the robot's actions to reach them.
Initiation of action is based on a hard trigger: that at least one
executable action exists that does not conflict with human actions.  In
contrast, in the work presented in this paper we aim to find a general
solution where acting is based on reasoning on first principles, rather than on
hard-coded triggers or rules.
\citet{bremner2019proactive} present a control architecture based on the
BDI-model incorporating an extra ethical layer in order to achieve
agents that are proactive, transparent, ethical and verifiable. They do
anticipation through embedded simulation of the robot and other
agents. Thereby, the robot can test what-if-hypotheses, e.g., what if I
carry out action $x$? The robot controller is given a set of goals,
task, and actions and thereof generates behavior alternatives, i.e.,
plans. The simulation module simulates them and predicts their
outcome. The ethical layer evaluates the plans and if needed invokes the
planner module to find new plans proactively.
Note that proactive plans are only generated if previously generated
plans from given goals fail against some given ethical rules, which
admittedly limits generality.  The approach that we propose below is more
general since it generates proactive actions from first principles.  On
the other hand, our approach does not takes ethics into account. \citet{umbrico2020holistic, umbrico2020toward} present a general-purpose
cognitive architecture with the aim to realize a Socially Assistive
Robot (SAR), specifically, for supporting elderly people in their home.
Their highly integrated framework includes a robot, a heterogeneous
environment and physiological sensors, and can do state assessment using
these sensors and an extensive ontology.  However, their approach to
making the SAR proactive is based on hard-wired rules like ``user need:
high blood pressure $\rightarrow$ robot action: blood pressure
monitoring in context \emph{sleeping}''. \citet{peng2019design} are mainly interested in finding the right level
of proactivity. They use hand-coded policies for guiding the behavior of
a robotic shopping assistant, and find that users prefer medium
proactivity over high or low proactivity.

\subsubsection{Equilibrium Maintenance} 

In this work we use Equilibrium Maintenance, $\eqm$, a mechanism
proposed by \citet{grosinger2019robots} for achieving proactivity.
$\eqm$ autonomously infers acting decisions based on temporal prediction
of one or several steps.  $\eqm$ is a general approach based on a formal
model, and thus affords domain independence.  This model is modular and
can cope with different agent capabilities, different preferences or
different predictive models.  The relation between situations and
triggered actions is not hard-coded: decisions are inferred at run-time
by coupling action with state, predicted states and preferences and
choosing among acting alternatives at run-time.  In
Section~\ref{ssec:eqm} we give a deeper description of $\eqm$.

A work that has comparable ideas to the ones in equilibrium maintenance
is the one on $\alpha$POMDPs by \citet{martins2019pomdp}. With the aim
to develop a technique for user-adaptive decision making in social
robots, they extend the classical POMDP formulation so that it
simultaneously maintains the user in valuable states and encourages the
robot to explore new states for learning the impact of its actions on
the user.  As in all flavors of (PO)MDPs, however, the overall objective
is to find an optimal, reward-maximizing policy for action selection; by
contrast, the aim of $\eqm$ is to maintain an overall desirable world
state, be it by acting or by inaction.  Instead of rewarding actions, as
done in MDPs, $\eqm$ evaluates the achieved effects of the actions (or
of being in-active).

\subsection{From Intention Recognition to Proactivity}

Several authors have proposed reactive systems based on intention
recognition.
\citet{Zhang2015} provide a framework for general proactive support in
human-robot teaming based on task decomposition, where priorities of
sub-tasks depend on the current situation. The robot re-prioritizes its
own goals to support humans according to recognized intentions.
Intentions are recognized by Bayesian inference following
\citet{ramirez2010}.  Each goal's probability depends on the agent's
past and/or current belief, and the goal with the highest probability
from a candidate goal set is recognized as the current intention.  Our
framework is similar to \citet{Zhang2015} in linking intention
recognition with the proactive behavior of the robot.  In our case,
however, the robot does not have its own independent tasks to achieve:
the robot's only objective is to help the human proactively by enacting
actions to reach their goal.

\citet{sirithunge2019proactive} provide a review of proactivity focused
on perception: robots perceive the situation and user intention by human
body language before approaching the human. The review aims to identify
cues and techniques to evaluate the suitability of proactive
interaction. Their idea of proactivity is that the robot identifies a
requirement by the human and acts immediately. This differs from our
understanding of reasoning on proactivity: we generate proactive agent
behavior by considering the overall environment, the human's intentions,
the overall preferences and prediction on how the state will
evolve. This can result in the agent acting now or later or not at all.

\citet{harman2020action} aim at predicting what action a human is likely
to perform next, based on previous actions observed through pervasive
sensors in a smart environment. Predictions can enable a robot to
proactively assist humans by autonomously executing an action on their
behalf.  So-called \emph{Action Graphs} are introduced to model order
constraints between actions. The program flow is as follows: (i)~action by the human is observed;
(ii)~next actions are predicted; (iii)~predicted actions are mapped to a
goal state; (iv)~a plan for the robot and a plan for the human are
created to reach the goal state; (v)~the robot decides which action it
should perform by comparing the robot's and the human's plan. The work presented in this paper shares some traits with the one by
\citet{harman2020action}: in both cases we reason on the human's
intentions, and make predictions about future states using action
models. In our case, however, predictions are made on how the system evolves
\emph{with} and \emph{without} robot actions, and proactive actions are
taken by comparing those predictions.  Importantly, in our case these
actions might \emph{not} be part of the human's plan.  Finally, the
trigger to perform proactivity reasoning in \citet{harman2020action} is
human action, while in our work this trigger is any state change, be it
caused by human action or by the environment.

In \citet{liu2021unified}, the authors' aim is to recognize and learn
human intentions online and provide robot assistance proactively in a
collaborative assembly task. They introduce the evolving hidden Markov
model (EHMM) which is a probabilistic model to unify human intention
inference and incremental human intention learning in real time.  \citet{liu2021unified} conduct experiments where a fixed robot arm
assists a human according to recognized intention in assembling cubes
marked by a fiducial mark on top. One such configuration corresponds to
one particular intention. The human starts to assemble the cubes and the robot proactively
finishes the shape as soon as the intention is recognized or when a
maximally probable intention is found by doing one step prediction. In \citet{liu2021unified} proactivity results from strict one-to-one
links where one recognized intention always leads to the same action
sequence, using a 1-step prediction.  In our approach, proactive robot behavior too can be based on recognizing intentions and their corresponding action sequences but it can also be inferred from first principles at run time using multiple steps
prediction.


\section{System}
\label{sec:sys}

\begin{figure}[thb]
\begin{center}
\includegraphics[ width=\textwidth,
  keepaspectratio]{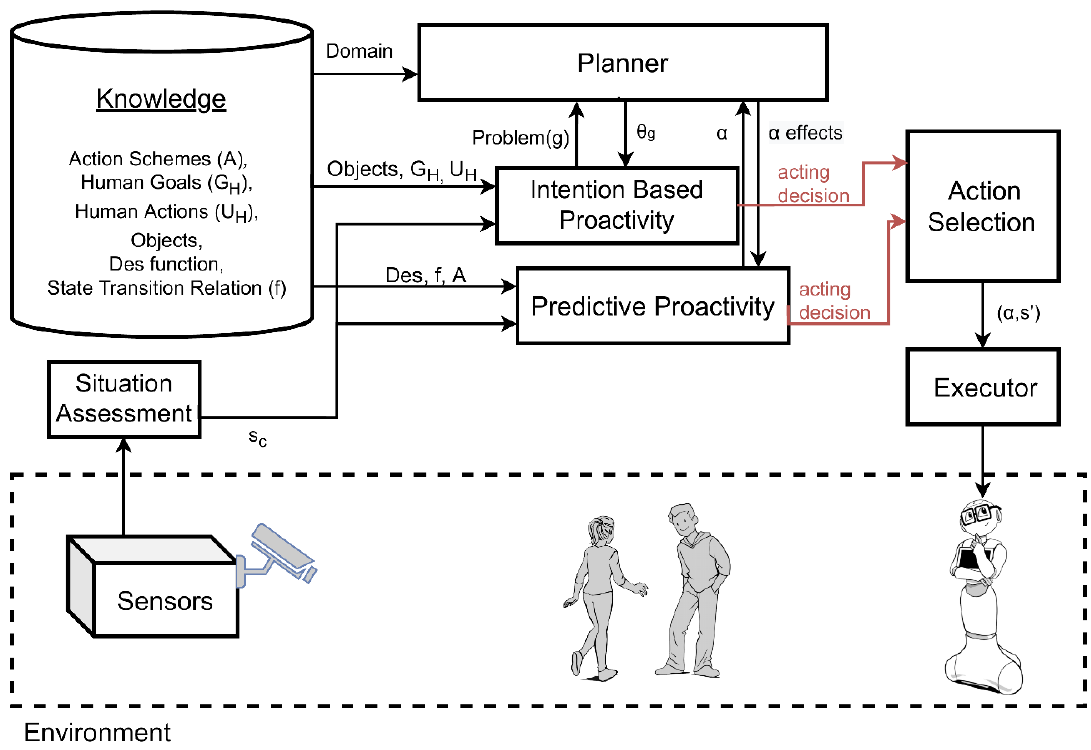}
\caption{System Model; an autonomous system that initiates proactive behavior according to the situation of the environment, including the human.
}\label{fig:system}
\end{center}
\end{figure}

We claimed that to initiate proactive behavior, robots must be equipped
with the abilities to recognize human intentions, to predict possible
future states and reason about their desirability, and to generate and
enact opportunities for acting that can lead to more desirable states.
In order to combine these abilities, we propose the general system model
shown in Figure~\ref{fig:system}.

The system includes different components to offer a fully autonomous
interaction, namely: a situation assessment, a knowledge component, a planner, an intention-based proactivity component, a predictive
proactivity component, an action selection component and lastly an
executor.
The \textbf{situation assessment} and the \textbf{executor} components
act as interfaces to the physical environment. They respectively collect
and induce changes from/to the environment.
The \textbf{knowledge} component represents a model of the
environment. This model encodes the state evolution of the world, the
set of goals of the human, action plans of how the human can reach their
goals, robot capabilities as a set of action schemes, the state transition relation and a desirability function to compute the
degree of desirability of a state.

More specifically, we model the environment and its dynamics using a
standard dynamic system $\Sigma = \langle S, U, f \rangle$ where $S$ is
a set of states, $U$ is a finite set of external inputs (robot actions
or human actions) and $f \subseteq S \times U \times S$ is a transition relation. The
system's dynamics is modeled by the relation $f(s,u,s')$, which holds
iff $\Sigma$ can go from $s$ to $s'$ when input $u$ is applied in $s$. 
To give $S$ a structure, we rely on a symbolic representation of world
states.  Given a finite set $\lang$ of predicates, we let $S \subseteq
\pset(\lang)$, and denote by $s_c$ the current state. Each state $s \in
S$ is thus completely determined by the predicates that are true in $s$.
We denote by $G_H \subseteq S$ the set of human goals. Each goal $g \in
G_H$ is determined by predicates that are true in $g$. Given a goal $g$,
we denote by $s_g$ any state in $S$ where all predicates in $g$ (and
potentially more) are true, hence $g \subseteq s_g$.  Finally, we denote
by $S_g \subseteq S$ the set of all states $s_g$ where the predicates of
$g$ are true.

The \textbf{planner}
is an off-the-shelf planner able to create a sequence of actions that
leads from the current state to a goal state. In our implementation, we
use \emph{Fastdownward}%
\footnote{\url{https://www.fast-downward.org/}},
a domain independent planner based on PDDL, the Planning Domain
Definition Language \citep{mcdermott_pddl_1998}. Both the human's plans and the robot's plans are formulated in PDDL, a standard language to
define planning domains and problems.  The planning \emph{domain}
includes the predicates of $\lang$ used for describing states, and
operators that model the available actions of humans and robots.  The
planning \emph{problem} includes information about the available
objects, the current state $s_c$, and the goal of the human $g \in G_H$.
Given a domain and a problem, the planner finds the shortest plan
$\theta_g(s)$ between the current state $s_c$ and the given goal
$g$. This plan represents the sequence of actions that the agent should
do to reach any state $s_g$ where all predicates of $g$ are true.

The \textbf{intention-based proactivity} component and the
\textbf{predictive proactivity} component are both able to generate
proactive behavior, but they use two different methods which we describe
below. Finally, the \textbf{action selection} component integrates the
decisions generated by those two methods into an overall proactive
behavior to be executed by the robot.

To describe the main contribution of this paper, i.e., the integration
of intention-based and predictive proactivity, we first need to
introduce the individual systems which it is based on.  In
Section~\ref{ssec:hir} we present our novel approach for intention-based
proactivity, $\uir$; In Section~\ref{ssec:eqm} we recall our existing
approach on equilibrium maintenance, $\eqm$; and in
Section~\ref{ssec:integr} we describe an integration of $\uir$ and
$\eqm$.

\subsection{Intention Based Proactivity: Human Intention Recognition and Reasoning}
\label{ssec:hir}

Experimental psychology shows that humans can interpret others'
intention by observing their actions, which is part of the so-called
Theory of Mind, ToM \citep{premack1978}.  Interpreting actions in terms
of their final goal may give hints on why a human performed those
actions, and hence make us able to infer that human's intentions
\citep{Han2013StateMaking}. Inspired from these concepts, we define a framework called \textit{Human Intention Recognition and Reasoning} ($\uir$) for
generating proactive behavior based on intention recognition.  
Intention recognition applies \emph{inverse planning} rules to recognize
the intention of the human in the form of an action plan. The robot can
then \emph{proactively} enact the next action in that action plan on
behalf of the human, or it can inform the human on which action to take
next in order to reach their goal.

There are different methods to recognize human intentions.  We select
inverse planning since this is a straight-through logic-based approach
for fully observable systems. The approach has been widely used in other
systems for intention recognition~\citep{Han2013StateMaking,
  persiani2020Computational, Farrell2020narrativeplanning}. While planning synthesizes a sequence of actions to reach a goal,
in inverse planning we observe the execution of a sequence of actions to infer the human's goal
and the corresponding plan.  Once the user has committed to reaching a
goal, we say the user \emph{intends} to reach that goal $g$.  We define
an intention $\inte(s)$ in state $s$ to be an action plan $\theta_g(s)$
to reach goal $g$ from state $s$. We infer human intentions $\ints(s)$ as defined in Equation~\ref{eq:intrec}:
\begin{equation}\label{eq:intrec}
    \ints(s) = \{\theta_{\hat{g}}(s) \mid \hat{g} \in \argmin_{g \in G_H}(\mathrm{len}(\theta_g(s))) \}
\end{equation}
In words, for each goal $g$ in the set $G_H$ of the human's potential
goals, we use our planner to compute the shortest plan $\theta_g(s)$
that the human can perform to reach $g$ from the current state $s$.  We
then select the goal $\hat{g}$ in $G_H$ to which the shortest of
these plans leads: $\theta_{\hat{g}}(s)$. The rationale behind this is that $\theta_{\hat{g}}(s)$ has the shortest
number of actions left to be executed, that is, the human already has
executed a large part of this plan.
Since we assume that the human is rational, it is plausible to infer
that the human intends to do all the remaining actions in
$\theta_{\hat{g}}(s)$ to reach $\hat{g}$ from $s$.  Therefore, we take
the action list in $\theta_{\hat{g}}(s)$ to be the intention $\inte(s)$
of the human in state $s$.
This strategy has been originally proposed in logic-based approaches in
\citet{persiani2020Computational}.

Equation~\ref{eq:intrec} is implemented by
Algorithm~\ref{alg:humtrick}, called $\uir$, that returns the intention
$\inte(s) \in \ints(s)$. The returned intention is the residual action plan $\theta_{\hat{g}}(s)$ of the
human's recognized intention to be enacted proactively by the robot.  If
the cardinality of the set of goals with shortest residual action  plans, i.e., the cardinality of the set of intentions, is not $1$, the intention is
not recognized or it is ambiguous and an empty set is returned.

%
%
\begin{algorithm}[H]
\dontprintsemicolon
\SetKwFunction{Return}{return}
\SetKwFunction{FirstAction}{first}
${\widehat{G}} = \argmin_{g \in G_H}(\mathrm{len}(\theta_g(s)))$\\
 \If{$\mid{\widehat{G}}\mid$ == 1}{
    $\hat{g} = \mbox{first}({\widehat{G}})$ \\
    \Return $\inte(s) = \theta_{\hat g}(s)$
 }\Else{
    \Return $\emptyset$ 
 }
\caption{$\uir(s, G_H)$\label{alg:humtrick}}
\end{algorithm}

\subsection{Predictive Proactivity: Equilibrium Maintenance}\label{ssec:eqm}

For doing reasoning on predictive proactivity we employ a computational
framework called \emph{Equilibrium Maintenance}, fully described in
\citet{grosinger2019robots}.  We only give a brief overview of the
framework here, the interested reader is referred to the cited reference
for details.

In our framework, the evolution of system $\Sigma$ by itself, that is,
when no robot action is performed, is modelled by its \emph{free-run
behavior} $\free{k}$.  $\free{k}$ determines the set of states that can
be reached from an initial state $s$ in $k$ steps when applying the null
input $\bot$.
\begin{align*}
\free{0}(s) &= \{s\} 
\\
\free{k}(s) &= \{s' \in S \mid \exists s'' :
   f(s, \bot, s'') \land s' \in \free{k-1}(s'') \}. 
\end{align*}
Desirable and undesirable states are modeled by $\des$, a fuzzy set of
$S$. The membership function $\ddes: S \rightarrow [0, 1]$ measures the
degree by which a state $s$ is desirable.  $\des$ is extended from
states to sets of states in the obvious way: $\ddes(X) = \tmin{s \in
  X}(\ddes(s))$, where $X \subseteq S$.  We abbreviate $\ddes(\cdot)$ as
$\des(\cdot)$.

The available robot actions are modeled by \emph{action schemes}: partial
functions
$  \alpha : \pset(S) \rightarrow \psetp(S) $
that describe how states can be transformed into other states by robot
acting.  An action scheme $\alpha$ abstracts all details of action:
$ \alpha(X) = Y $ 
only says that there is a way to go from any state in the set of states
$X$ to some state in set $Y$.  Action schemes can be at any level of
abstraction, from simple actions that can be executed directly, to
sequential action plans, or policies, or high level goals for one or
multiple planners.  Applying an action scheme $\alpha$ in a state $s$
may bring about effects that are (or are not) desirable, possibly in $k$
steps in the future.  We call \emph{benefit} the degree to which
an applied action scheme achieves desirable effects:
\begin{align}\label{eq:bnf}
\bnf(\alpha,s,k) = \tmin{X \in \dom(\alpha,s)} \des(\free{k}(\alpha(X))),
\end{align}
where
$\free{k}(X) = \bigcup_{s \in X}\free{k}(s)$ and $\dom(\alpha,s)$ is the
domain of $\alpha$ relevant in $s$.   

With this background, \citet{grosinger2019robots} define seven different
types of \emph{opportunity} for acting, which are the foundation of
proactivity by equilibrium maintenance.  We write $\opp_i(\alpha,s,k)$
to mean that applying action scheme $\alpha$ in state $s$ is an
opportunity of type $i$, by looking $k$ steps into the future.
\begin{align*}
\opp_0(\alpha,s,0) &= \min(1 - \des(s), \bnf(\alpha,s))\\
\opp_1(\alpha,s,k) &= \min(1 - \des(s), \tmax{s' \in \free{k}(s)}(\bnf(\alpha,s')))\\
\opp_2(\alpha,s,k) &= \min(1 - \des(s), \tmin{s' \in \free{k}(s)}(\bnf(\alpha,s'))) \\
\opp_3(\alpha,s,k) &= \tmax{s' \in \free{k}(s)}(\min(1 - \des(s'), \bnf(\alpha,s'))) \\
\opp_4(\alpha,s,k) &= \tmin{s' \in \free{k}(s)}(\min(1 - \des(s'), \bnf(\alpha,s'))) \\
\opp_5(\alpha,s,k) &= \min(\tmax{s' \in \free{k}(s)}(1 - \des(s')), \bnf(\alpha,s,k)) \\
\opp_6(\alpha,s,k) &= \min(\tmin{s' \in \free{k}(s)}(1 - \des(s')), \bnf(\alpha,s,k))
\end{align*}
To understand these opportunity types, consider for example the first
type $\opp_0$: the degree by which $\alpha$ is an opportunity of type
$0$ is the minimum of (i) the degree by which the current state $s$ is
undesirable, and (ii) the benefit of acting now.  Intuitively, $\alpha$
is an opportunity of type $0$ if (and to the extent) we are in an
undesirable state, but enacting $\alpha$ would bring us to a desirable
one.  As another example, consider $\opp_5$: here, we compute the
minimum of (i) the maximum undesirability of future states, and (ii) the
future benefit of acting now: intuitively, $\alpha$ is an opportunity of
type $5$ if (and to the extent) some future states within a look-ahead
$k$ are undesirable, but if we enact $\alpha$ now then all the $k$-steps
future states will be desirable.

Finally, we can define what it means for a system to be in equilibrium
from a proactivity perspective.
\begin{align}\label{eq:eq}
\eq(s,K) = 1 - \tmax{k,i,\alpha}\opp_i(\alpha,s,k),
\end{align}
where $k \in [0, K]$, $i \in [0, 6]$, $\alpha \in A$, where $A$ is
the set of all action schemes. 
Intuitively, equilibrium is a measure of lack of opportunities: if there
are big opportunities, then the system is very much out of equilibrium;
if there are small opportunities, then the system is close to being in
equilibrium; if there are no opportunities at all, then the system is
fully in equilibrium. The notion of equilibrium is used in the
equilibrium maintenance algorithm $\eqm$ to achieve agent proactivity,
as shown in Algorithm~\ref{alg:eqmaintm}.%
\footnote{This algorithm is a slightly modified version of the original
version given in~\citet{grosinger2019robots}; this is done to
accommodate for the integration with human intention recognition and
reasoning, $\uir$.}

%
\begin{algorithm}[H]
\dontprintsemicolon
\SetKw{KwTrue}{true}
\SetKwFunction{Return}{return}
\SetKwFunction{Choose}{choose}
\If{$\eq(s,K) < 1$}{
  $\opps \leftarrow \arg\max_{k,i,\alpha}(\opp_i(\alpha,s,k))$ \; 
  $\langle \alpha, s',\opp_i, k, \oppdeg \rangle \leftarrow $\Choose($\opps$) \;
  \Return $\langle \alpha, s',\opp_i, k, \oppdeg \rangle$
} \Else{
    \Return $\emptyset$ 
}
\caption{$\eqm(s, K)$\label{alg:eqmaintm}}
\end{algorithm}
%

\subsection{Action Selection: Integrating Human Intention Recognition
  and Reasoning and Equilibrium Maintenance}
\label{ssec:integr}

$\uir$ and $\eqm$ are complementary approaches that create proactive
acting in different ways.  We now explore how to integrate the two
systems.  The action selection component in Figure~\ref{fig:system}
integrates the approaches at the result phase, after each system has
proposed their proactive action.  However, each approach has a different
reasoning mechanism and affects the future states in different ways.
$\uir$ supports the human towards reaching their intentions.  It infers
the human's intention and suggests, or enacts, a sequence of actions to
reach the human's goal starting from the current state.
$\eqm$ prevents the human from being in undesirable states by predicting
possible state evolutions, and reasoning on what is desirable and how
available robot actions could create benefit.

Integrating the two systems is not trivial. 
Consider the hiking example in the opening of this paper, and suppose
that in a given state $s$, $\eqm$ infers an opportunity to warn the human
for hail, $\opp_{5}(\alpha_{\mathrm{warn}}, s, 2)$.  Suppose that at the
same time $\uir$ recognized that the human intention is to go hiking and
infers to bring the compass to the human.
 
We have two competing goals for robot acting, and action selection needs
to weigh them via a common scale. We propose a solution for integrating
$\eqm$ and $\uir$ by turning the goal from $\uir$ into an opportunity of
type $\opp_0$, hence, $\opp_{0}(\alpha_{\mathrm{collect(compass)}}, s,
0)$ and check its degree. In other words, we check the desirability of the states that would be
achieved by the action when applied.  Note that we use $\opp_0$ here
since the decisions by $\uir$ are meant to be acted upon immediately and
do not use multiple step lookahead, just like $\opp_0$.  Once we have
converted the individual outputs from $\uir$ and from $\eqm$ to a common
format, that is, sets of opportunities, we collect all these
opportunities into a pool, from which the Action Selection component
(Figure~\ref{fig:system}) chooses an acting alternative.

To transform an $\uir$ acting decision into an opportunity of type
$\opp_0$ we temporarily modify the outcome of the $\des$-function. In the $\eqm$ framework, the $\des$-function does not take human
intentions into account. It does not model states with unfulfilled human
intentions as undesirable and those with fulfilled ones as
desirable. Such a $\des$-function would therefore not generate an
opportunity corresponding to an unfulfilled intention. We therefore
temporarily modify $\des$ to decrease the desirability of the current
state (Algorithm~\ref{alg:uir}, line 3), modeling the undesirability of
unfulfilled intention, and increase the desirability of the effects of
an action that fullfills the intention (line 4): this allows the
generation of an opportunity based on human intention recognition. For example, a state that would be desirable to the degree $0.7$ by
itself might only be $0.1$-desirable when a certain human intention has
been recognized.  Conversely, we increase the desirability of the
effects that would manifest when an action of the human's intention is applied, which
would not be the case otherwise when no human intention was recognized,
e.g., with recognized human intention $\des(\alpha(s)) = 0.9$, without
recognized human intention $\des(\alpha(s)) = 0.3$.

\begin{algorithm}[H]
\dontprintsemicolon
\SetKw{KwTrue}{true}
\SetKwFunction{Increase}{increase}
\SetKwFunction{Decrease}{decrease}
\SetKwFunction{Return}{return}
\SetKwFunction{First}{first}

  $\alpha(s) \leftarrow$ \First($\uir (s, G_H)$) \; 
  
  \If{$(\alpha \neq \bot)$}{
    
    $\desp(s) \leftarrow$ \Decrease ($\des(s)$)
    
    $\desp(\alpha(s)) \leftarrow$ \Increase ($\des(\alpha(s))$)
  
    
    $\oppdeg \leftarrow \min(1 - \desp(s), \bnfp(\alpha,s))$ 
  
  \Return $\langle \alpha, s,\opp_0, 0, \oppdeg \rangle$ 
  }\Else{
  \Return $\langle \rangle$
  }
\caption{$\uir\!\text{-Opp}(s, G_H)$}
\label{alg:uir}
\end{algorithm}

In our experiments, we have implemented the \texttt{decrease} and
\texttt{increase} functions by scaling by a fixed value; exploring
better ways to implement these steps is a matter for further
investigation. The modified desirability function $\des'(s)$ is used in line 8 to
compute the degree of the opportunity of type $\opp_0$ for applying
action scheme $\alpha$ in the current state $s$.  $\alpha$ is the first
action in the recognized intention (action plan) returned by $\uir$
(line 1). Note that the computation in line 5 uses a modified
$\bnf'(\alpha, s)$, which is based on $\des'(\alpha(s))$. This opportunity, and its degree, are returned in line 6, and represent
the opportunity based on human intention recognition.

Now that we have opportunities for acting based on prediction, as
returned from Algorithm~\ref{alg:eqmaintm}, and the one based on
reasoning on human intention, as returned from Algorithm~\ref{alg:uir},
we can decide which of them to enact in using the \emph{action
selection} Algorithm~\ref{alg:oppcomb}.
This algorithm continuously checks if the state has changed (line 4), be it by changes in the
environment or by application of robot action.  If so,
it collects the opportunities coming from both proactivity systems,
$\eqm(s,K)$ and $\uir(s,G_H)$ (lines 5 and 6) and then chooses one of
these to be dispatched to the executive layer and enacted (line 8). The function $\Choose()$, like the $\Choose()$ in
Algorithm~\ref{alg:eqmaintm}, can implement several strategies. In our experiments, $\Choose()$ selects the opportunity with the highest
degree to be enacted. If there are several opportunities with highest
degree a decision is made by the opportunity type, how much benefit can
be achieved and the size of the look-ahead.  More discussion on these
strategies can be found in
\citep{grosinger2019robots}.

%
\begin{algorithm}[H]
\dontprintsemicolon
\SetKw{KwTrue}{true}
\SetKwFunction{Return}{return}
\SetKwFunction{Choose}{choose}
\SetKwFunction{Add}{add}
\SetKwFunction{Dispatch}{dispatch}

\While {\KwTrue} {
  $s \leftarrow$ current state \;
  $\opps \leftarrow \{\}$\;
  \If{$s$ has changed }{
    $\opps.$\Add($\uir\!\text{-Opp}(s, G_H)$) \;
    $\opps.$\Add($\eqm(s, K)$)\;
    $\langle \alpha, s',\opp_i, k, \oppdeg \rangle \leftarrow $\Choose($\opps$) \;    
   
    \Dispatch($\alpha, s'$)\;
    }
}
\caption{Action Selection ($K, G_H$)
\label{alg:oppcomb}}
\end{algorithm}
%
%
\section{Experiments}\label{sec:exp}

In this section, we empirically explore the behavior of the presented
approaches to proactive behavior by running the same simulated task with
three different configurations of the system.  We compare and analyze
the outcomes of the system shown in Figure~\ref{fig:system} when:

\begin{enumerate}
    \item we only use intention based proactivity ($\uir$);
    \item we only use predictive proactivity ($\eqm$);
    \item we integrate both intention based and predictive proactivity.
\end{enumerate}

\begin{figure}[tbh!]
  \centering
  \begin{minipage}[b]{0.49\textwidth}
    \includegraphics[width = \textwidth]{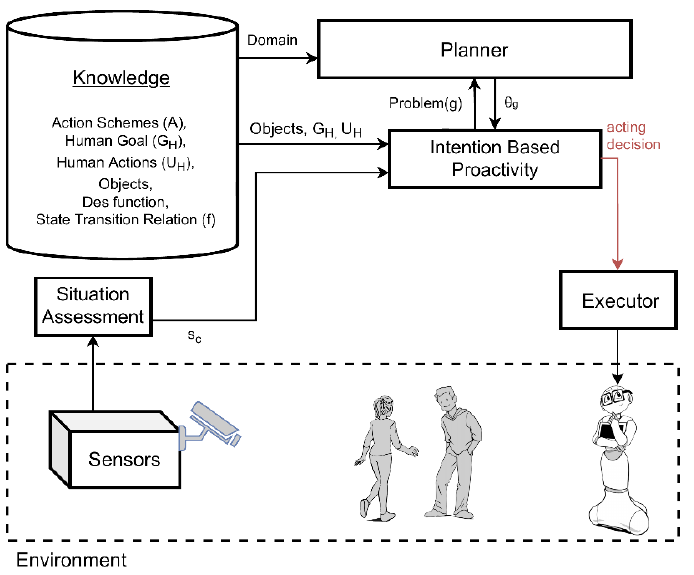}
    \caption{System model overview for $\uir$}
  \label{fig:system_hir}
  \end{minipage}
  \hfill
  \begin{minipage}[b]{0.49\textwidth}
    \includegraphics[width = \textwidth]{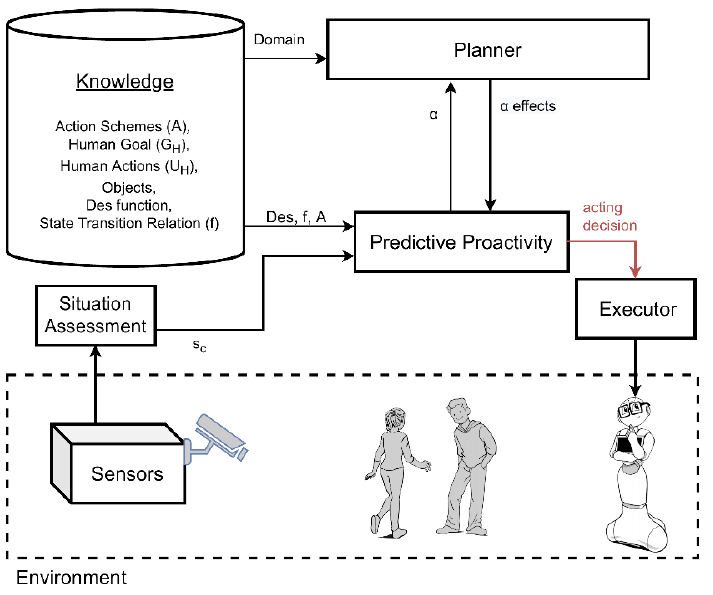}
    \caption{System model overview for $\eqm$}
\label{fig:system_eqm}
  \end{minipage}
\end{figure}

Figures~\ref{fig:system_hir} and Figure~\ref{fig:system_eqm} show the
systems used for the first two experiments, while the integrated system
used for the third experiment is the one previously shown in
Figure~\ref{fig:system}.

The code is available in a ``research bundle'' on the European
AI-on-demand platform, \url{ai4europe.eu}%
\footnote{\url{https://www.ai4europe.eu/research/research-bundles/proactive-communication-social-robots}}.
This research bundle includes the open source code, libraries, and a
form of a notebook allowing users to interact with the framework by
defining their environment.

\subsection{Task Description}

\begin{figure}[htbp]
\begin{center}
\includegraphics[width=\textwidth, keepaspectratio]
  {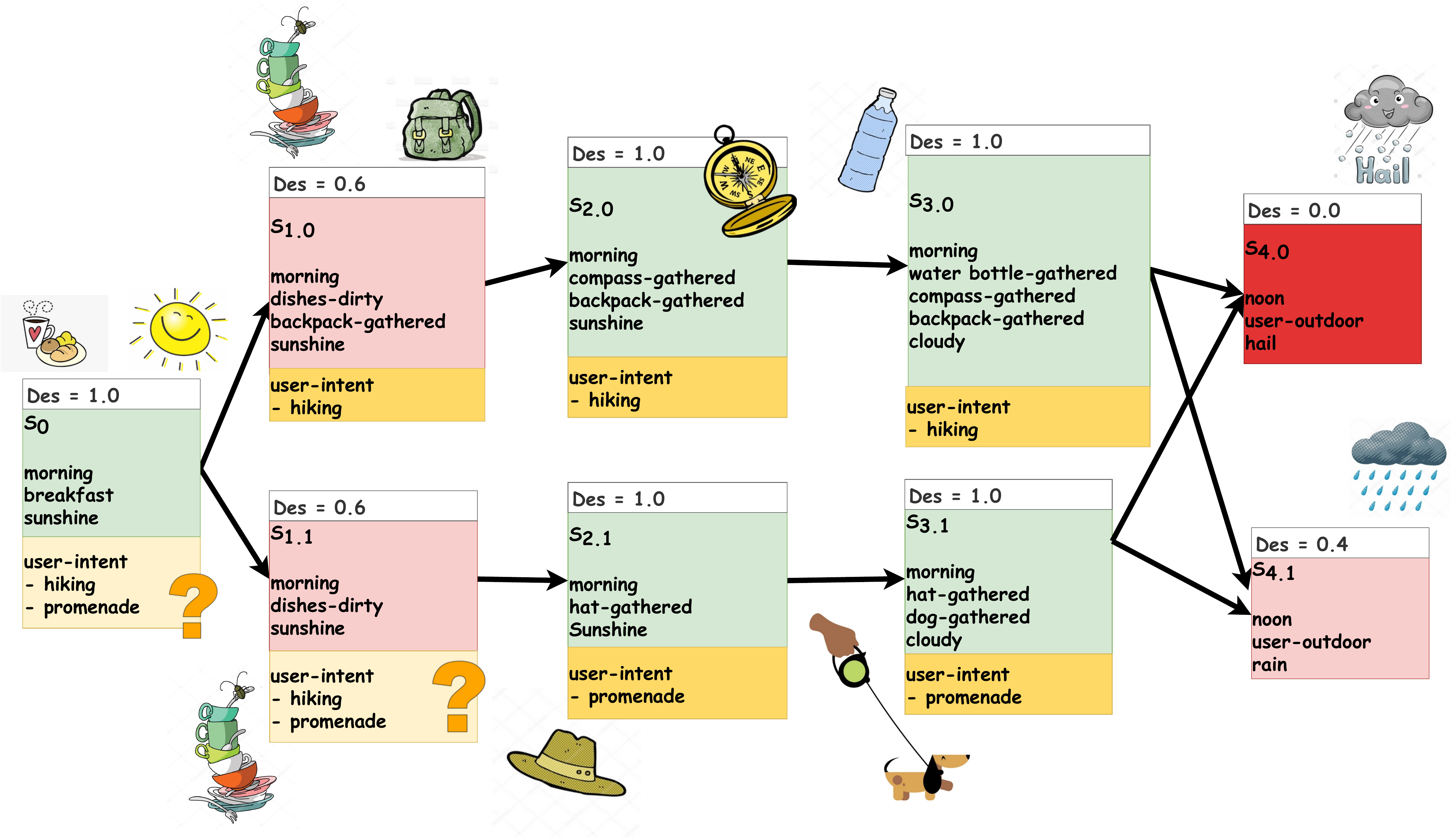} 
\caption{States and possible state transitions (free-run) in our
  scenario.  Desirability values for each state are color-coded, as well
  as indicated numerically.  Green represents desirable states, while
  pink to red represents less desirable states. The more undesirable a
  state is, the more intense its red tone. }
\label{fig:stateev}
\end{center}
\end{figure}

We define a hypothetical scenario where a human moves inside a house
and collects objects in order to reach a goal. Figure ~\ref{fig:stateev}
graphically represents the dynamic system $\Sigma = \langle S, U, f
\rangle$ that models our scenario, where arrows show the state
transitions $f$ that correspond to possible evolutions of the
environment (including actions by the human) if there is no interference
from the robot.  The figure also indicates the degree of desirability
$\des(s)$ of each state $s$.
In addition to $\Sigma$ and $\des$, the scenario includes the set of
action schemes given in Table~\ref{tab:actions} below, and a set $G_H$
of four human goals:
\begin{itemize}
    \item \textbf{Hiking;} backpack collected, compass collected, water
      bottle collected, human is outside.
    \item \textbf{Promenade;} hat collected, dog collected, walking
      stick collected, human is outside. 
    \item \textbf{Watch TV;} water bottle collected, sugar collected,
      tea collected, remote control collected. 
    \item \textbf{Read Book;} glasses collected, book collected, tea
      collected, sugar collected. 
\end{itemize}
%
Each goal describes what must be true for it to be considered
reached. For example, the goal ``Hiking'' is reached when it is true
that backpack, compass and water bottle are collected and the human is
outside. The actions to reach the individual goals can be done both by
the human or the robot (except for going outside).

In our implementation, both the dynamic system and the action schemes
are modelled in PDDL.  Recall that PDDL includes a domain definition and
a problem definition. In the domain definition, we model object
definitions, predicate definitions for logical facts, and action
definitions with preconditions and effects.  In the problem definition,
we model the initial state of the environment in logical format, as well
as the goal state. Actions are defined for \textit{gathering} and \textit{leaving objects},
for \textit{telling the human they are ready to leave the house}, and
for \textit{cleaning the dishes}. Details are given in Table~\ref{tab:actions}. Some actions can be
executed by the robot, some can be executed by the human, and some by
both the human and the robot. The actions that are done by the human are
observed by the robot, and based on them the $\uir$ system recognizes
the human's intention. The $\eqm$ system, on the other hand, reasons
about potential robot actions while taking into account the human's
actions which are part of the free-run (uncontrollable state
transitions). Note that in this use case all actions are deterministic
except for \emph{cleaning the dishes} which is non-deterministic: the
action can have the effect that all dishes are clean or that they are
still half dirty.

The defined robot actions are used in the $\eqm$ system to infer
opportunities.  $\uir$ recognizes human intention by inferring the
human's action plan. When the intention is recognized, $\uir$ can make
the robot proactively carry out the rest of the human's action plan on
the human's behalf. However, the human's action plan towards their goal
might contain an action that cannot be carried out by the robot. In that
case the robot transforms the action to a communication action where the
robot tells the human what they should do. For example, after having
collected all the necessary items, the human is supposed to leave the
house to reach their goal ``hiking''. The robot can collect all
necessary items but cannot leave the house, hence, it tells the human
``Everything has been collected. You are ready to leave now for going
hiking''.
\begin{table}[h]
    \centering
    \begin{tabular}{|l l l r|}
    \hline
        \textbf{Action} & \textbf{Precondition} & \textbf{Effect} & \textbf{Agent} \\ \hline \hline 
         Gather object & \textbf{and (} & human gathered object & human / robot \\ 
          & obj is not gathered,  &  &  \\ 
          &  human at home \textbf{)} &  & \\ \hline
         Leave object & \textbf{and (} & human not gathered object & human / robot \\ 
          & obj is gathered, & &  \\ 
         & human at home \textbf{)} &  &  \\ \hline
         Leave home & human at home & human not at home & human \\
         \hline
        Suggest human to leave home & human at home & human not at home & robot \\
         \hline
        Warn human & human at home & \textbf{and (}  & robot \\
        & & human at home, & \\
        & &  human warned \textbf{)} & \\
         \hline
        Clean dishes  & \textbf{or (} & \textbf{or (}  & human / robot \\
         & dishes dirty,  & dishes not dirty & \\
         & dishes half dirty \textbf{)}  & dishes half dirty ) & \\
         \hline
    \end{tabular}
    \caption{The table provides the actions that the human and the robot
      are capable to do. It provides the name of the action (possibly
      including a parameter), preconditions of the actions and the
      effects that will show after the action is applied, as well as
      who can do the action (human and/or robot).} 
    \label{tab:actions}
\end{table}
The desirability function, $\des$, that computes the desirability degree
of each state, is assumed given. In our example scenario, we consider one specific run where the state
evolves as follows: $s_{0}$ , $s_{1.0}$ , $s_{2.0}$ , $s_{3.0}$ (see
Figure~\ref{fig:stateev}). The system starts in $s_{0}$ where the
weather is nice, time is morning and the human is having breakfast. This
state is very desirable, $\des(s_{0}) = 1.0$. Later the state is
changed to $s_{1.0}$ where the weather is still nice and the time is
still morning, but the human finished their breakfast so there are dirty
dishes and the human collected the backpack. This state is less
desirable, $\des(s_{1.0}) = 0.6$. Later the state evolves to $s_{2.0}$,
where the weather is cloudy and dishes are cleaned. In addition to the
backpack, now the compass is collected. The last state chance is to
$s_{3.0}$, where the weather is cloudy, time is morning and the human
has collected the water bottle in addition to the previously collected
belongings backpack and compass. Note that the predicate
``dishes-dirty'' changes from true (in $s_{1.0}$ and $s_{1.1}$) to false
(in $s_{2.0}$ and $s_{2.1}$). This is because the free-run state
evolution models all uncontrollable state transitions which includes the
environment and the human. Hence, the dishes not being dirty anymore
means that the human has taken care of cleaning them.

\subsection{Human Intention Recognition and Reasoning Only}
\label{ssec:hironly}

We consider the scenario described in Figure~\ref{fig:stateev} using
only human intention recognition and reasoning for achieving proactive
agent activity. This means we evaluate an implementation of the method
$\uir$ as described in Section~\ref{ssec:hir}. The architecture of the
system is shown in Figure~\ref{fig:system_hir}. Table~\ref{tab:hir}
lists the recognized human intentions in the respective state and the
proactive agent activity inferred.
\begin{table}[h!]
    \centering
    \begin{tabular}{|l|l|l|}
    \hline
        \textbf{State} & \textbf{Intention recognized} & \textbf{Proactive agent activity chosen - $\uir$} \\
        \hline \hline
         $s_0$ & ? & --- \\
         \hline
         $s_{1.0}$ & hiking & gather water bottle \\
         \hline
         $s_{2.0}'$ & hiking & tell human that he/she is ready to leave the house \\
         \hline
         $s_{3.0}$ & ? & --- \\
         \hline

    \end{tabular}
    \caption{The state evolution and the proactive agent activity inferred in each state when using human intention recognition and reasoning only.}
    \label{tab:hir}
\end{table}
In $s_0$, the $\uir$ cannot recognize yet what the human's intention is,
it could be any of the four known human goals, going on a hike or going
on a promenade, watching TV or reading a book. 
Then the state advances to $s_{1.0}$ where a backpack is collected. In
this state, the $\uir$ is able to detect that the human's intention is
to go on a hike. The $\uir$ can infer to \emph{proactively} bring the
water bottle to the human as this is the next action inferred in the
human's action plan.
The action is dispatched and the robot proactively brings the water
bottle to the human (in simulation). Now the human has the backpack
(gathered by the human him-/herself) and the water bottle (gathered by
the robot). The state evolution advances to the next state
$s_{2.0}'$. In this state also a compass is gathered, which was done by
the human. 
(For any state $s$ in
the free-run in Figure~\ref{fig:stateev}, $s'$ marks its equivalent on
which robot action has been applied.)
Again, the intent recognition detects that the human's intention is
going on a hike. The intention reasoning system detects that all
necessary items for going on a hike have been collected. Therefore, it
infers the \emph{proactive} activity of notifying the human that he/she
is ready to leave.  
Note that, in state $s_{2.0}'$ (which is the state $s_{2.0}$ plus
applied robot action) the same predicates are true as in $s_{3.0}$. This
is eligible and expected as the robot's proactive acting is doing part
of the human's action plan based on human intention recognition and
reasoning. Therefore, a state $s_{2.0}'$ would not evolve into $s_{3.0}$
(which is identical), but into states $s_{4.0}$ or $s_{4.1}$ where the
human is outdoors.
Note that, once the human has left the house, proactive interaction from
our system with him/her is not possible. That is why there is no
intention recognized in $s_{4.0}, s_{4.1}$.  
Note also that these states are quite undesirable ($\des(s_{4.0}) = 0.0$
and $\des(s_{4.1}) = 0.4$), as the user is outdoors while weather
conditions are unpleasant (\texttt{rain}) or even dangerous
(\texttt{hail}).
The algorithm for human intention recognition and reasoning neither does
any prediction of future states nor reasons about
desirability/preference. Therefore, it is ignorant of the upcoming
undesirable situation and cannot act on it.

\subsection{Equilibrium Maintenance Only}
We again consider the use case described in Figure~\ref{fig:stateev} but
now using equilibrium maintenance only for achieving proactive agent
activity. This means we test an implementation of the  equilibrium
maintenance algorithm as described in
Section~\ref{ssec:eqm}. Table~\ref{tab:eqm} lists the opportunities for
acting inferred in the respective state and the proactive agent activity
to be enacted, i.e., the chosen opportunity.  
%
%
The outcome of $\eqm$ depends very much on the size of the prediction,
$K$. In our system run, we set $K = 2$.
(See \citet{grosinger2019robots} for a discussion on the choice of the
look-ahead horizon $K$.)
%
We let $\eqm$ infer opportunities for acting that are current ($k = 0$),
one time step in the future ($k = 1$), and two time steps in the future
($k = 2$). 

The state evolution starts in state $s_0$.  The following opportunities
are inferred: $\eqm$ does not infer to act in the current state, $s_0$,
because the human is having breakfast and the state is very desirable,
$\des(s_{0}) = 1.0$. However, when projecting the state 1 time step into
the future, $\eqm$ observes that the upcoming possible states will be
less desirable, $\des(s_{1.0}) = 0.6$ and $\des(s_{1.1}) = 0.6$, because
there will be dirty dishes from the breakfast. Therefore, $\eqm$ infers
the opportunities for the robot to put the dishes in the dishwasher in
the future, i.e., 1 step from now. There are no opportunities for acting
in 2 time steps.  
%
When the human gathers the backpack, state $s_0$ evolves to $s_{1.0}$. The state is not very desirable, $\des(s_{1.0}) = 0.6$, because there are dirty dishes from the breakfast. $\eqm$ infers the opportunity for the robot to put the dirty dishes in the dish washer now, in the current state $s_{1.0}$. Note that, what before in $s_{0}$ had been an opportunity for acting in the future, is now an opportunity for acting in the present. 
\begin{table}[h!]
    \centering
 \begin{tabular}{|l|l|l|}
    \hline 
        \textbf{State} & \textbf{Opportunities inferred} & \textbf{Proactive agent activity chosen} \\
        \hline \hline
         $s_0$ 
        &
        $\opp_{3,4,5,6}(\alpha_{\mathrm{gather(any)}}, s, 1) = 0.01$  & \\
        &  \fcolorbox{llgray}{llgray}{$\opp_{3,4}(\alpha_{\mathrm{clean}}, s, 1) = 0.4$ } & Clean dishes in 1 step\\

        \hline
         $s_{1.0}$ 
          &
        $\opp_{0}(\alpha_{\mathrm{gather(any)}}, s, 0) = 0.01$  & \\
         &  \fcolorbox{llgray}{llgray}{$\opp_{0}(\alpha_{\mathrm{clean}}, s, 0) = 0.4$ } & Clean dishes now\\
         & $\opp_{1,2}(\alpha_{\mathrm{gather(any)}}, s, 1) = 0.01$  & \\
        & $\opp_{1,2}(\alpha_{\mathrm{gather(any)}}, s, 2) = 0.01$  & \\
         \hline
         
        $s_{2.0}$ 
       
         &  $\opp_{5,6}(\alpha_{\mathrm{gather(any)}}, s, 2) = 0.01$  & \\
        & 
        \fcolorbox{llgray}{llgray}{$\opp_{5}(\alpha_{\mathrm{warn}}, s, 2) = 1.0$ } & Warn for hail, effect seen in 2 steps\\
       &  $\opp_{6}(\alpha_{\mathrm{warn}}, s, 2) = 0.6$  & \\
        \hline
        
        $s_{3.0}$ 
         &
        $\opp_{5,6}(\alpha_{\mathrm{gather(any)}}, s, 1) = 0.01$  & \\
        &  \fcolorbox{llgray}{llgray}{$\opp_{5}(\alpha_{\mathrm{warn}}, s, 1) = 1.0$ } & Warn for hail, effect seen in 1 step\\
       & $\opp_{6}(\alpha_{\mathrm{warn}}, s, 1) = 0.6$  & \\
        \hline
\end{tabular}
    \caption{The state evolution and the proactive agent activity inferred in each state when using equilibrium maintenance only. Note that $\alpha_{\mathrm{warn}}$ refers to warning the human for risk of bad/harmful weather conditions, $\alpha_{\mathrm{clean}}$ refers to cleaning the dishes, i.e., putting them in the dishwasher, $\alpha_{\mathrm{gather(any)}}$ refers to gather any object for the human.}
    \label{tab:eqm}
\end{table}
When the human gathers the compass, the state evolves into $s_{2.0}$. The state is desirable again, $\des(s_{2.0}) = 1.0$, because the dishes are not dirty any more. 
This means either the robot has enacted the opportunity of putting the dishes in the dish washer in the previous state, or putting the dishes in the dish washer has happened through uncontrollable action as part of the free-run, i.e., the human has put the dishes in the dish washer. In $s_{2.0}$ there are no opportunities considering the current state or taking a look-ahead of $k = 1$. However, when $\eqm$ projects the state two steps into the future, it appears that the possible states are very undesirable, $\des(s_{4.0}) = 0.0$, and quite undesirable, $\des(s_{4.1}) = 0.4$. This is because the human will be outdoors and the weather will be very bad (rain) or even dangerous (hail). $\eqm$ therefore infers an opportunity for the robot to act now in order to prevent the future very undesirable outcome. More concretely, the robot \emph{proactively} goes to the human and warns the human now in order to prevent him/her to be outdoors in the hail later. In case the warning is not heeded, in $s_{3.0}$ the same opportunity for acting is inferred, only that now the look-ahead is one time step instead of two.
%


\subsection{Human Intention Recognition and Reasoning and Equilibrium Maintenance}
As before, we consider the use case described in Figure~\ref{fig:stateev} but now using both $\uir$ and $\eqm$ for achieving proactive agent activity. This means we have a system as described in Section~\ref{ssec:integr}. 
Table~\ref{tab:hireqm} lists the opportunities inferred by $\uir$  and the opportunities inferred by $\eqm$, in the respective state and marks which of them is chosen to be enacted. 
In $s_0$, there are no opportunities by $\uir$ since the human intention cannot be unambiguously determined yet which yields zero opportunities for proactive acting (see Section~\ref{ssec:hir}).
$\eqm$, on the other hand, does infer opportunities for acting in state $s_{0}$. The one opportunity with the greatest degree, and hence, the one chosen, is an opportunity to clean the dishes in 1 time step from now, $\opp_{3,4}(\alpha_{\mathrm{clean}}, s_{0}, 1)$. 
%
\begin{table}[h!]
\centering
\resizebox{\textwidth}{!}{
\begin{tabular}{|l|l|l|m{11em}|}
    \hline 
        \textbf{State} & \textbf{Proactive acting --- $\uir$} & \textbf{Proactive acting  --- $\eqm$} & \textbf{Chosen Proactive Action} \\
        \hline \hline

     $s_0$ 
        & &  $\opp_{3,4,5,6}(\alpha_{\mathrm{gather(any)}}, s, 1) = 0.01$  & \\
       & &  \fcolorbox{llgray}{llgray}{$\opp_{3,4}(\alpha_{\mathrm{clean}}, s, 1) = 0.4 $}  & Clean dishes in 1 step \\
     
        \hline
         $s_{1.0}$ 
         & &
        $\opp_{0}(\alpha_{\mathrm{gather(any)}}, s, 0) = 0.01$  & \\
       & \fcolorbox{llgray}{llgray}{ $\opp_{0}(\alpha_{\mathrm{gather(wb)}}, s, 0) = 0.5$} &  $\opp_{0}(\alpha_{\mathrm{clean}}, s, 0) = 0.4$ & Gather water bottle now \\
        & &
        $\opp_{1,2}(\alpha_{\mathrm{gather(any)}}, s, 1) = 0.01$  & \\
                & &
        $\opp_{1,2}(\alpha_{\mathrm{gather(any)}}, s, 2) = 0.01$  & \\
         \hline
    
        $s_{2.0}$ 
         & $\opp_{0}(\alpha_{\mathrm{leave}}, s, 0) = 0.5$ &  $\opp_{5,6}(\alpha_{\mathrm{gather(any)}}, s, 2) = 0.01$  & \\
        &  &  \fcolorbox{llgray}{llgray}{$\opp_{5}(\alpha_{\mathrm{warn}}, s, 2) = 1.0$ } & Warn for hail, \\
           &  &  $\opp_{6}(\alpha_{\mathrm{warn}}, s, 2) = 0.6$  & effect seen in 2 steps \\
        \hline
        $s_{3.0}$ 
         & $\opp_{0}(\alpha_{\mathrm{leave}}, s, 0) = 0.5$ &
        $\opp_{5,6}(\alpha_{\mathrm{gather(any)}}, s, 1) = 0.01$  & \\
         & &  \fcolorbox{llgray}{llgray}{$\opp_{5}(\alpha_{\mathrm{warn}}, s, 1) = 1.0$ } & Warn for hail, effect seen in 1 step \\
       & & $\opp_{6}(\alpha_{\mathrm{warn}}, s, 1) = 0.6$  & \\
        \hline
\end{tabular}
}
    \caption{The state evolution and the proactive agent activity inferred in each state when using both human intention recognition and reasoning and equilibrium maintenance. (Note that 
    $\alpha_{\mathrm{warn}}$ refers to warning the human for risk of bad/harmful weather conditions, $\alpha_{\mathrm{clean}}$ refers to cleaning the dishes, i.e., putting the dishes in the dish washer, $\alpha_{\mathrm{gather(any)}}$ refers to gathering any object for the human, and $\alpha_{\mathrm{leave}}$ refers to informing the human that he/she is ready to leave the house.)
    }
    \label{tab:hireqm}
\end{table}
In $s_{1.0}$, there is an opportunity coming from $\uir$ since the human intention is recognized as "hiking". Hence, there is an opportunity to gather the water bottle, $\opp_0(\alpha_{\mathrm{gather(wb)}},s_{1.0},0)$. The degree and the type of the opportunity is computed according to Algorithm~\ref{alg:uir}. 
Also $\eqm$ produces opportunities in $s_{1.0}$. The opportunity from $\eqm$ which has the highest degree is cleaning the dishes now --- this is the opportunity from $s_0$, now being of type $\opp_0$ (to be applied now), while it was an opportunity for the future, $\opp_{3,4}$, in $s_0$. $\uir$'s opportunity to gather the water bottle now is chosen to be enacted. How to choose between the opportunities from $\uir$ and $\eqm$ is determined by Algorithm~\ref{alg:oppcomb}. 

The human now has the backpack (gathered by the human him-/herself) and the water bottle (gathered by the robot). The state evolution advances to the next state $s_{2.0}'$. Note that, in state $s_{2.0}'$ (which is the state $s_{2.0}$ plus applied robot action) the same predicates are true as in $s_{3.0}$ (see also Section~\ref{ssec:hironly}).
%
$\uir$ recognizes going on a hike as the intention of the human and proposes the opportunity to inform the human that he/she is ready to leave the house as all belongings for the hike have been packed. 
$\eqm$ proposes opportunities for warning the human about undesirable (rain) or possibly dangerous weather conditions (hail) as the human is predicted to be outdoors in the future, two time steps from now. The opportunity coming from $\eqm$ (warning the human) has a higher degree, $0.6$, than the opportunity coming from $\uir$ ("ready-to-go"-message for the human) which has a degree of $0.5$. Therefore, the combined system of $\uir$ and $\eqm$ chooses to dispatch the robot activity of warning the human. 

In $s_{3.0}$, since the conditions have not changed 
$\uir$ infers the opportunity to confirm the human that they have gathered all necessary items and hence are ready to leave the house for going hiking. The warning of $\eqm$ has not been heeded in the previous state by the human, hence, $\eqm$ again infers to warn the human for the future unpleasant/dangerous weather, only now just one time step into the future instead of two. 

\section{Discussion and Conclusion}\label{sec:disc}

In this paper, we have analyzed two approaches to proactivity: $\uir$,
human intention recognition and reasoning, that infers proactive actions
by recognizing the human's intended plans and taking over the next
action in these plans; and $\eqm$, equilibrium maintenance, that infers
opportunities for acting by reasoning about possible future states and
about what states are preferable. 
We have then defined a third approach which
combines these two types of proactivity, and have illustrated the three
approaches on a sample use case.
%
%
Our analysis shows that each approach can generate some proactive
behaviors but not others. $\uir$ focuses on helping humans towards
achieving their intentions, whereas $\eqm$ focuses on preventing humans
to end up in undesirable situations. 
$\eqm$ does not
consider humans' intentions and therefore cannot generate proactive
behavior to support the human to achieve them; $\uir$, on the other hand, does not
reason about how the state will evolve and about the overall
desirability of future states, and therefore cannot generate proactive
behavior based on the predicted benefit of actions.
The combined system can take into account both, humans' intentions and the
desirability of future states.  In this work, we have explored a rather
simple way to combine these two aspects, where $\uir$ and $\eqm$
independently propose proactive actions and one of those is selected.
This can be called \emph{late integration}.  Future work will
investigate \emph{early} forms of integration, where reasoning on human
intentions, on available robot's actions, on future states and on
preferences among those states is done in an integrated fashion.  This
tighter integration of $\uir$ and $\eqm$ will require a shared
formulation for the two.

Our current framework is built on state descriptions that only consider
the physical world.  In future work, it will be interesting to include
the inner world of the human.  We plan to explore the use of techniques
from the area of epistemic logic and epistemic planning to model
intentions, knowledge and beliefs of human agents.  The $\des$-function
and benefit of acting can then also take into account the preferences of
mental states and how to bring preferable epistemic states about.

While we considered a single human in this paper, both $\uir$ and $\eqm$
can in principle consider multiple humans: one would then have to track
separately the single actions of each person and infer their intentions.
If the humans are collaborating, $\uir$ could consider all of the
humans' actions together to infer the collective intention. $\eqm$ can
fuse the single humans' preferences in one overall $\des$-function.

The results of $\eqm$, as well as the results of the combined system
$\uir$ + $\eqm$, strongly depend on the models of the dynamics of the
system $\Sigma$, which determines the prediction of the state evolution
(free-run), and on the modeling of preferences ($\des$-function).
To see this, consider the example in Section~\ref{sec:exp}. In state
$s_{1.0}$ the opportunity of $\uir$, gather the water bottle, is chosen
over the opportunity of $\eqm$, clean the dishes.  If the desirability
function modeled a stronger undesirability of dirty dishes, then the
degree of the opportunity by $\eqm$ would have been higher and hence
this opportunity would have been chosen to be enacted.
Further, consider a slightly modified state dynamics, that differs from
the one in Figure~\ref{fig:stateev}. In these modified state dynamics, the states at time $4$ have weather conditions 'sun' and 'clouds' (instead of 'rain' and 'hail'). In this use case, $\eqm$ would only
infer to clean the dishes in state $s_{1.0}$ but it would not support
the human preparing for their hike because it is not reasoning on the
human's intention. $\uir$, on the other hand, could provide a lot of
support to the human to achieve their intention and would achieve a very
desirable final outcome: the human out on a hike with all their
belongings gathered.
%

In this paper, we have simply assumed that both $\Sigma$ and $\des$ are
given.  Future work, however, might explore the use of machine learning
to learn probabilistic models of state evolution. Also hybrid techniques
(model-based and data-driven) are conceivable which should take into
account inferring the human's intentions as this is an indicator of what
the human will do next and thereby how the state will change.  When
eliciting and reasoning on preferences, ideally, the proactive agent
should take into account the dynamic change of preferences, weigh
personal against common, long-term against short-term preferences, and
reason on uncertainty of what is desirable.


Finally, the example in Section~\ref{sec:exp} has been run with console
inputs and outputs.  Given the promising results, the next step will be
to connect our system to a real Pepper robot
(\citep{Pandey2018AKind}) and conduct a user study in a real domestic
environment.
Since our system is based on general models, we also plan to test it on
a variety of physical robot systems or interactive agents in diverse domains.

\section*{Author Contributions}


SB and JG contributed equally to the concept, formalization,
implementation and experimentation reported in this paper. MC and AS
supervised all aspects above.  All four authors contributed equally the
planning, writing and revision of the paper.


\section*{Acknowledgments}

This  project  has  received  funding from European Union’s  Horizon 2020 ICT-48 research and innovation actions under grant agreement No 952026 (HumanE-AI-Net) and from  the  European Union’s  Horizon 2020 research  and  innovation programme under grant agreement No 765955 (ANIMATAS).

\bibliographystyle{frontiersinSCNS_ENG_HUMS}
\bibliography{refs}

\end{document}